\documentclass[runningheads]{llncs}
\usepackage[T1]{fontenc}
\usepackage{graphicx}
\usepackage{booktabs}
\usepackage[misc]{ifsym}
\newcommand{\corr}{(\Letter)}
\usepackage{mwe}

\usepackage{algorithm}
\usepackage{algorithmicx}
\usepackage{amsmath,amsfonts}
\usepackage{mathabx}
\usepackage[noend]{algpseudocode}
\usepackage{textcomp}
\usepackage[caption=false]{subfig}
\usepackage[dvipsnames]{xcolor}
\usepackage{graphicx}
\usepackage{bbm}
\usepackage{multirow}

\DeclareMathOperator*{\argmin}{arg\,min}

\title{Conformalized Exceptional Model Mining: Telling Where Your Model Performs (Not) Well}
\toctitle{Conformalized Exceptional Model Mining: Telling Where Your Model Performs (Not) Well}

\titlerunning{Conformalized Exceptional Model Mining}

\author{Xin Du\inst{1,2,3,4}  \and
Sikun Yang\inst{1,2,3,4}\corr \and
Wouter Duivesteijn \inst{5} \and
Mykola Pechenizkiy \inst{5}
}
\tocauthor{Xin Du, Sikun Yang, Wouter Duivesteijn, Mykola Pechenizkiy}
\authorrunning{X. Du et al.}

\institute{School of Computing and Information Technology, Great Bay University, 523000 Dongguan, China \email{\{duxin,sikunyang\}@gbu.edu.cn}
\and
Great Bay Institute for Advanced Study, Great Bay University
\and
Guangdong Provincial Key Laboratory of Mathematical and Neural Dynamical Systems, Great Bay University
\and
Dongguan Key Laboratory for Intelligence and Information Technology, Great Bay University
\and
Data and Artificial Intelligence Cluster, Department of Mathematics and Computer Science, Eindhoven University of Technology, 5600MB Eindhoven, the Netherlands \email{\{w.duivesteijn, m.pechenizkiy\}@tue.nl}
}

\begin{document}

\maketitle

\begin{abstract}
Understanding the nuanced performance of machine learning models is essential for responsible deployment, especially in high-stakes domains like healthcare and finance. This paper introduces a novel framework, Conformalized Exceptional Model Mining, which combines the rigor of Conformal Prediction with the explanatory power of Exceptional Model Mining (EMM). The proposed framework identifies cohesive subgroups within data where model performance deviates exceptionally, highlighting regions of both high confidence and high uncertainty. We develop a new model class, mSMoPE (multiplex Soft Model Performance Evaluation), which quantifies uncertainty through conformal prediction's rigorous coverage guarantees. By defining a new quality measure, Relative Average Uncertainty Loss (RAUL), our framework isolates subgroups with exceptional performance patterns in multi-class classification and regression tasks. Experimental results across diverse datasets demonstrate the framework's effectiveness in uncovering interpretable subgroups that provide critical insights into model behavior. This work lays the groundwork for enhancing model interpretability and reliability, advancing the state-of-the-art in explainable AI and uncertainty quantification.

\keywords{Exceptional Model Mining \and Conformal Prediction \and Uncertainty Quantification \and Explainability}

\end{abstract}

\section{Introduction}
As deep learning techniques continue to advance across various fields such as healthcare~\cite{ahmad2018interpretable} and financial data science~\cite{provost2013data}, it becomes increasingly important to ensure that these models are both responsible and explainable when deployed~\cite{arrieta2020explainable}. For example, in the case of a deep learning model used for disease diagnosis, medical professionals may need to understand the conditions under which the model provides highly confident predictions, as well as situations where it exhibits significant uncertainty. To meet these demands, uncertainty quantification methods are essential~\cite{shafer2008tutorial}, while techniques such as Exceptional Model Mining (EMM) can offer interpretable insights~\cite{du2020uncertainty,duivesteijn2016exceptional,leman2008exceptional}. Therefore, developing a conformalized exceptional model mining approach is crucial for helping users gain a deeper understanding of their model's performance~\cite{smith2024uncertainty}.

EMM focuses on modeling multivariate interactions (captured in a \emph{model class}), and discovering cohesive subgroups where these interactions are exceptional compared with these interactions across the whole dataset~\cite{duivesteijn2016exceptional}. EMM has been widely applied to the analysis of classification, regression tasks for tabular datasets~\cite{duivesteijn2012different}, structural interaction for the network datasets~\cite{kaytoue2017exceptional}, and transition behavior for sequential datasets~\cite{lemmerich2016mining}. EMM practitioners have applied this framework on more complex datasets and diverse scenarios, including explainable machine learning research where EMM is employed to analyze the behavior of classifiers to provide responsible and explainable results~\cite{proencca2022robust}. However, existing classification model classes in EMM only focus on the binary classification problem: application for regression and multi-class classification (especially for the softmax based classifiers) is still underexplored. We develop a new model class for EMM to understand the behavior of regressors and multi-class classifiers.

Conformal prediction is a versatile framework designed to provide rigorous guarantees for uncertainty quantification~\cite{angelopoulosuncertainty}. Given a heuristic probability estimate from any pre-trained model ---whether for classification, regression, or other tasks—-- conformal prediction leverages a small hold-out dataset to transform this heuristic estimate into a statistically valid probability measure with guaranteed coverage. Specifically, in a classification scenario, conformal prediction generates a prediction set that contains the true label with a certain confidence level. This prediction set serves as a measure of uncertainty, offering insights into the classifier's reliability. A smaller prediction set indicates higher confidence in the model's predictions, while a larger set reflects greater uncertainty.

We strive to generate explainable results that identify cohesive subgroups and provide insights into the model's performance, through a method that can be applied to any classifier or regressor. To do so, we must address two major difficulties. On the one hand, due to the multi-output nature of multi-class classifiers using the softmax function, it is crucial to design a model class capable of handling complex interactions between multiple variables. On the other hand, integrating the validity in conformal prediction within concepts of interestingness within EMM presents another significant challenge. To address these issues, we propose a novel model class called mSMoPE (multiplex Soft Model Performance Evaluation), which introduces the average size of uncertainty sets as a new target variable for the EMM framework. This allows input attributes to be leveraged in forming cohesive subgroups that capture the relationship between the attributes and the target variable.

In this work, we utilize conformal prediction across various tasks, including classification and regression on tabular data~\cite{angelopoulos2023prediction}. Based on these applications, we introduce a general framework that demonstrates how conformalized exceptional models can be applied to diverse data mining scenarios to uncover meaningful cohesive patterns. These patterns, represented as conjunctions of attribute-value conditions, offer valuable insights into the behavior of machine learning models. Furthermore, by incorporating deep learning models such as multilayer perceptrons (MLPs), our framework enables a deeper understanding of deep learning models for tabular datasets, facilitating comprehensive performance analysis in tabular tasks.

\subsection{Main Contributions}
We propose a general framework for exceptional model mining with conformal prediction named conformalized exceptional model mining. This framework allows the user to generate a rigorous output about the model's performance with uncertainty measurement. Our main contributions are:

\begin{itemize}
  \item we present mSMoPE: multiplex Soft Model Performance Evaluation, a new model class for EMM. This model class processes the input attributes, distilling a rigorous uncertainty set that captures the uncertainty of the model's output; 
  \item based on the proposed new model class, we define a new quality measure that measures the average uncertainty loss between subgroups and the entire dataset. The proposed quality measure allows us to represent the discrepancy quantitatively so that the exceptional interplay between the input attributes and the uncertainty set can be revealed;
  \item we conduct experiments qualitatively and quantitatively on several public datasets. The results effectively demonstrate the interpretable subgroups that are exceptional in terms of predictive uncertainty.
\end{itemize}

\section{Related Work}

Exceptional Model Mining (EMM) \cite{du2020uncertainty,duivesteijn2016exceptional,leman2008exceptional} is a subfield of pattern mining, itself a branch of data mining, that focuses on identifying patterns in subsets of data rather than the dataset as a whole. Pattern mining is distinct in that it seeks to describe only specific portions of the data, often with a predefined description language, while ignoring the coherence of the remaining data. A subset of the data is considered ``interesting'' based on some criteria of interest, and EMM builds on this principle. In pattern mining, such subsets are often described using a conjunction of conditions on dataset attributes. For instance, in a dataset where records describe people, a pattern might look like ``Age $\geq$ 30 AND Smoker = yes $\rightarrow$ interesting''. By restricting patterns to conditions that relate to attributes of the data, the results become more interpretable to domain experts, as they align with familiar quantities. Such subsets, expressed in terms of these conditions, are called \emph{subgroups}.

One of the best-known forms of pattern mining is Frequent Itemset Mining \cite{1994Agrawal}, which identifies subgroups that co-occur unusually often in an unsupervised manner: ``Age $\geq$ 30 AND Smoker = yes $\rightarrow$ high frequency''. In supervised settings, the focus shifts to identifying subgroups based on their relationship to a specified target attribute. This is the foundation of Subgroup Discovery (SD) \cite{2015Atzmueller,2011Herrera,1996Kloesgen,1997Wrobel}, which aims to find subgroups where the distribution of a binary target attribute is unusual: ``Smoker = yes $\rightarrow$ Lung cancer = yes''. EMM extends Subgroup Discovery by considering multiple target attributes simultaneously. Rather than focusing on unusual distributions of a single target attribute, EMM investigates unusual interactions between several target attributes.
EMM achieves this by defining a \emph{model class} to represent the type of unusual interaction between targets, and a quality measure to quantify the interestingness of subgroups. The aim is to find subgroups where this quality measure is maximized, thereby uncovering patterns that exhibit exceptional relationships within the data.

Existing work has explored the identification of subgroups exhibiting unusual interactions among multiple targets, through established EMM model classes such as correlation, regression, Bayesian networks, and classification~\cite{leman2008exceptional,duivesteijn2016exceptional,duivesteijn2010subgroup}. Among these, the classification model class \cite[Section 3.3]{leman2008exceptional} bears particular relevance to the mSMoPE model class, though they differ in two key aspects. First, their definitions establish distinct relationships between subgroup descriptions and the classifier's search space. In the classification model class, both the input and output attributes of the classifier are treated as targets, preventing these attributes from appearing in subgroup descriptions; exceptional subgroups are thus described using attributes excluded from the classifier. In contrast, the mSMoPE model class permits all input attributes (excluding outputs) to define subgroups, directly linking discovered subgroups to subspaces within the classifier's search space. Second, the model classes pursue different goals: the classification model class analyzes classifier behavior without a ground truth, whereas the mSMoPE model class evaluates performance in the presence of a ground truth during the calibration stage. 
The closest work to this paper is SCaPE~\cite{duivesteijn2014understanding}. This model class studies soft classifiers for a binary target, seeking subgroups of the classifier input space where the soft classifier outputs are exceptionally well or badly aligned with the binary ground truth. Conversely, in this paper, we focus on the performance of a multi-class classifier or a regressor. We construct the target variable based on the concept of conformal prediction, for which the terms of uncertainty have been included in the target.

Several Local Pattern Mining tasks share similarities with SD, including Contrast Set Mining~\cite{bay2001detecting} and Emerging Pattern Mining~\cite{dong1999efficient}. These tasks do not address multiple target attributes simultaneously, and do not explicitly model unusual interactions. Distribution Rules~\cite{jorge2006distribution} represent an approach where a deviating model over a single numeric target is sought, identifying subgroups where the target’s distribution deviates most from the overall dataset distribution. While this can be viewed as an early form of EMM with a single target, it does not account for multi-target interactions. In contrast, Umek et al.~\cite{umek2011subgroup} consider SD with multiple targets, but their method reverses the attribute partitioning approach used in EMM. They generate candidate subgroups through agglomerative clustering of the targets and use predictive modeling on the descriptors to find matching subgroup descriptions. However, this approach does not allow for flexible expressions of unusual target interactions. Redescription Mining~\cite{galbrun2012black} identifies multiple descriptions that induce the same subgroup, modeling unusual interactions within the descriptor space rather than the target space. None of these methods explicitly evaluate the performance of a classifier or a regressor.

Many studies focus on accurately estimating predictive uncertainty in neural networks. Initially, the standard approach involved training Bayesian neural networks to learn a distribution over network weights, which required both computational and algorithmic modifications~\cite{gal2016uncertainty,kuleshov2018accurate}. Alternative methods circumvent these challenges through ensembles~\cite{lakshminarayanan2017simple} or approximate Bayesian inference~\cite{sensoy2018evidential}. However, these approaches have limitations, such as the need to train multiple neural network copies adversarially. Consequently, the most commonly used technique remains the ad-hoc calibration of softmax scores using Platt scaling~\cite{guo2017calibration}.
Conformal prediction offers a different perspective by generating predictive sets that satisfy the coverage property~\cite{vovk2005algorithmic}. We employ a practical data-splitting variant known as split conformal prediction, which enables the application of conformal prediction techniques to virtually any predictor~\cite{lei2018distribution}. Unlike traditional calibration methods, conformal prediction operates within a general framework rather than a specific algorithm. Therefore, key design choices must be made to optimize performance for different contexts.
In this paper, our primary contribution is the integration of the conformal prediction framework into EMM, equipping the latter with the ability to quantify prediction confidence.

\section{Conformalized Exceptional Model Mining}
\label{sec:CEMM}

We assume a dataset $\Omega=(X,Y)$, where $X$ stems from a $k$-dimensional \emph{input space} $\mathcal{X}=\bigtimes_{i=1}^k\mathcal{X}^i$, and $Y$ stems from an \emph{output space} $\mathcal{Y}$, to be instantiated later. We also assume a \emph{predictor} $\mu:\mathcal{X}\to\mathcal{Y}$. We denote the number of records in $\Omega$ by $N$, and we allow each $\mathcal{X}^i$ to have any reasonable domain: binary, categorical, ordinal, numeric. We partition the $N$ records of $\Omega$ into two parts: the \emph{calibration set} $\Omega_{\text{calib}}=(X_{\text{calib}},Y_{\text{calib}})$ and the \emph{test set} $\Omega_{\text{test}}=(X_{\text{test}},Y_{\text{test}})$. We denote the number of records in $\Omega_{\text{calib}}$ by $n$, leaving $N-n$ records for $\Omega_{\text{test}}$.

\subsection{Conformal Prediction}

Conformal Prediction (CP) is a distribution-free framework in machine learning and statistical modeling that provides valid confidence estimates, prediction sets, or prediction intervals for predictive models. In this work, we focus on the computationally efficient split CP method~\cite{vovk2005algorithmic}. It performs as a wrapper around a trained base model and uses a set of exchangeable hold-out / calibration data to construct prediction sets or intervals. Given a predefined miscoverage rate $\alpha \in [0,1]$, the method follows three main steps. Firstly, it computes non-conformality scores, which quantify the degree to which a given output deviates from expected predictions. CP defines a non-conformality score function $S: \mathcal{X} \times \mathcal{Y} \rightarrow \mathbb{R}$, which captures uncertainty in the model’s predictions. Intuitively, $S(x, y)$ measures how $y$ ``conforms'' to the prediction at x; in classification, this could be the predicted probability of class $y$, while in regression, it may be the residual value $S(x, y) = |y - \hat{\mu}(x)|$ for a predictor $\hat{\mu}: \mathcal{X} \rightarrow \mathcal{Y}$. Secondly, it computes the $(1 - \alpha)$ quantile of the non-conformality scores on the calibration set. For $\Omega_{\text{calib}}=\{(X_i, Y_i)\}^n_{i=1}$, $\hat{q}$ can be computed as $\text{Quantile}(\{S(X_1, Y_1), \ldots, S(X_n, Y_n)\}, [(1-\alpha)(n+1)]/n)$. Thirdly, for any $(X_{n+1},Y_{n+1})\in\Omega_{\text{test}}$, Conformal Prediction constructs a prediction set or interval $\mathcal{T}(X_{n+1}) = \{y \in \mathcal{Y} : S(X_{n+1}, y) \leq \hat{q}\}$. Additionally, if $\{X_i, Y_i\}^{n+1}_{i=1}$ are exchangeable, then $S_{n+1} := S(X_{n+1}, Y_{n+1})$ is exchangeable with $\{S_i\}_{i=1}^{n}$ since $\hat{\mu}$ is given. Hence, $\hat{\mathcal{T}}(X_{n+1})$ contains the true label with predefined coverage rate~\cite{vovk2005algorithmic}: $P\{Y_{n+1} \in \mathcal{T}(X_{n+1})\} = \mathbb{P}\{S_{n+1} \geq \text{Quantile}(\{S_1, \ldots, S_{n+1}\}, 1-\alpha)\} \geq 1 - \alpha$ due to exchangeability of $\{S_{i=1}^{n+1}\}$. This framework works with any non-conformality score; we discuss one for multi-class classification, and one for regression.

\subsubsection{Adaptive Prediction Set (APS)} 

In classification tasks, general methods generating prediction sets usually create smallest average sizes, but tend to undercover hard subgroups and overcover easy ones. Adaptive Prediction Set (APS) is proposed to avoid this problem; we use the non-conformality score in APS proposed by~\cite{romano2020classification}.  The cumulative sum of ordered class probabilities was taken till the true class. As motivation for this procedure, note that if the softmax outputs $\hat{\mu}(X_{\text{test}})$ were a perfect model of $Y_{\text{test}} | X_{\text{test}}$, we would greedily include the top-scoring classes until their total mass exceeds $1 - \alpha$. Formally, we can describe this oracle algorithm as $\{\pi_1, \dots, \pi_k\}$, where $k = \inf \{k^{\prime}:\sum_{j=1}^{k^{\prime}} \hat{\mu}(X_{\text{test}})_{\pi_j} \geq 1 - \alpha \}$, and $\pi$ is the permutation of $\{1, \ldots, K\}$ that sorts $\hat{\mu}(X_{\text{test}})$ from most likely to least likely. Because we never know whether $\hat{\mu}(X_{\text{test}})$ is any good, this procedure fails to provide coverage. Hence, we need to use CP to transform the heuristic notion of uncertainty into a rigorous notion. We define a score function as $s(x, y) = \sum^k_{j = 1}\hat{\mu}(x)_{\pi_j}$, where $y = \pi_k$. That is to say, we greedily include classes in our set until we reach the true label, then we stop. This procedure considers the softmax outputs of all classes rather than just the true class. Then we conduct the conformal procedure to set $\hat{q} = \text{Quantile}(\{S_1, \ldots, S_n\}, [(1-\alpha)(n+1)]/n)$, and formulate the prediction set as $\mathcal{T}(x) = \{\pi_1, \ldots, \pi_k\}$, where $k = \inf \{k^{\prime}: \sum^{k^{\prime}}_{j=1} \hat{\mu}(x)_{\pi_j} \geq \hat{q} \}$.

\subsubsection{Conformalized Quantile Regression (CQR)} 

Conformalized Quantile Regression (CQR)~\cite{romano2019conformalized} is a widely recognized CP method for constructing prediction intervals, known for its simplicity and effectiveness. CQR is based on quantile regression that acquires heuristic estimates $\hat{\mu}_{\alpha/2}(x)$ and $\hat{\mu}_{1-\alpha/2}(x)$ for the $(\alpha/2)$ and $(1 - \alpha/2)$ conditional quantile functions of $Y$ given $X = x$. The non-conformality score is computed using the calibration set as: $S_i = \max \{\hat{\mu}_{\alpha/2}(X_i) - Y_i, Y_i - \hat{\mu}_{1-\alpha/2}(X_i) \}$, for each $(X_i, Y_i) \in \Omega_{\text{calib}}$. Then the scores are employed to calibrate the plug-in prediction interval $\hat{\mathcal{T}}(x) = [\hat{\mu}_{\alpha/2}(x), \hat{\mu}_{1 - \alpha/2}(x)]$. To be specific, we let $\hat{q}$ be the $([(|\Omega_{\text{calib}}| + 1)(1-\alpha)] / |\Omega_{\text{calib}}|)$ empirical quantile of $\{S(X_1, Y_1), \ldots, S(X_{|\Omega_{\text{calib}}|}, Y_{|\Omega_{\text{calib}}|})\}$; the prediction interval for new input data $X_{\text{test}}$ is then constructed as $\mathcal{T}(X_{\text{test}}) = [\hat{\mu}_{\alpha/2}(X_{\text{test}}) - \hat{q}, \hat{\mu}_{1 - \alpha/2}(X_{\text{test}}) + \hat{q}]$.

\subsection{Soft and Hard Model Outputs}

Suppose that we have a multi-class classification problem: any record of the dataset belongs to exactly one of the $K$ available classes. Let us denote those classes by $\{1, \ldots, K\}$. For such a problem, two particular types of classification algorithms can be distinguished. On the one hand, a hard classifier outputs for each record in the test set a decision to which class it thinks the record belongs: the output is one of the $K$ values. On the other hand, a soft classifier outputs for each record in the test set a real-valued $K-$plex vector (typically but not necessarily a probability): the output can be any value in $\mathbb{R}$, and higher values for the output correspond to a higher confidence that the records should be assigned class corresponding to the indicator. 

Next, suppose that we have a real-valued regression problem: any record of the dataset could be mapped to exactly real-valued numbers. Let us denote those numbers by $Y$. For such a problem, two particular types of regression algorithms can be distinguished. On the one hand, a hard regressor outputs for each record in the test set a decision to which number it thinks the record can be mapped to: the output is a real-valued number. On the other hand, a soft regressor outputs for each record in the test set an interval with which one represents the lower quantile in the conditional distribution function of Y, and the other represents the upper quantile. 
Fix the lower and upper quantiles to be equal to $\alpha_{\text{lo}} = \alpha / 2$ and $\alpha_{\text{hi}} = 1 - \alpha / 2$. We will have an interval that covers the true value $Y$ with miscoverage rate $\alpha$, with the lower and upper pair of conditional quantile functions as $q_{\alpha_{\text{lo}}}(x)$ and $q_{\alpha_{\text{hi}}}(x)$, as $\mathcal{T}(x) = [q_{\alpha_{\text{lo}}}(x), q_{\alpha_{\text{hi}}}(x)]$. By construction, this interval satisfies $\mathbb{P}\{Y \in \mathcal{T}(X)|X=x\} \geq 1 - \alpha$. 
The length of the prediction interval $\mathcal{T}(x)$ can vary substantially depending on the value of $X$. This variation naturally captures the uncertainty in predicting $Y$, with wider intervals indicating greater uncertainty~\cite{meinshausen2006quantile,takeuchi2006nonparametric}.

Hard regressors estimate the test response $Y_{n+1}$ given the record $r^{n+1} = x$ by minimizing the sum of squared residuals on the $n$ training points: $\hat{\mu}(x) = \mu(x, \hat{\theta})$, $\hat{\theta} = \argmin_{\theta} \frac{1}{n} \sum_{i=1}^{n} (Y_i - \mu(X_i;\theta))^2 + \mathcal{R}(\theta)$. Here $\theta$ are the parameters of the regression model and $\mu(x; \theta)$ is the regression function, and $\mathcal{R}$ is a potential regularizer. Conversely, soft regressors like quantile regression estimate a conditional quantile function $q_{\alpha}$ of $Y_{n+1}$ given $X_{n+1} = x$. This can be cast as the optimization problem $\hat{q}_{\alpha}(x) = f(x;\hat{\theta})$, $\hat{\theta} = \argmin_{\theta}\frac{1}{n} \sum_{i=1}^{n} \rho_{\alpha}(Y_i, f(X;\theta)) + \mathcal{R}(\theta)$, where $f(x;\theta)$ is the quantile regression function and the loss function $\rho_{\alpha}$ is the ``pinball loss'' defined by 

\[ \rho_{\alpha}(y, \hat{y}):=
  \begin{cases}
    \alpha(y - \hat{y}) & \quad \text{if } y-\hat{y} > 0,      \\
    (1-\alpha)(\hat{y} - y)  & \quad \text{otherwise.}
  \end{cases}
\]
This makes quantile regression widely applicable. In a concrete setting with a dataset at hand, we have information about each record $r^i$, whose true label or true response can be used as the perfect output of the hard model output. At the same time, we can also extract input attribute information to compute the soft model output. We investigate unusual interplay between these classifiers and regressors in an Exceptional Model Mining setting.

\subsection{The mSMoPE Model Class for EMM}\label{sec:msmope}

Traditional EMM assumes a dataset $\Omega$, which is a bag of $N$ records of the form $(a_1, \ldots, a_k, r)$. Candidate subgroups are generated by a guided search through the space spanned by the \emph{descriptors} $a_1,\ldots,a_k$, and generated candidates are evaluated by a quality measure $\varphi$ that assesses the candidate for exceptional behavior on the target space $r$; see \cite[Algorithm 1]{duivesteijn2016exceptional} for an algorithm performing this task. As $k$ EMM descriptors, we can simply employ the $k$-dimensional input space as introduced near the start of Section \ref{sec:CEMM}. The remaining challenge is to extract a useful target space $r$ out of the conformal predictions in the preceding sections, and define a quality measure $\varphi$ that sensibly determines whether generated candidate subgroups display exceptional conformal predictive behavior.

The goal of the mSMoPE model class for EMM is to let $r$ capture the size of prediction set or the interval of regressing prediction, which represent the uncertainty level of how the model performs, and seek subgroups where this uncertainty is extreme: one can parameterize the model class to seek highly certain subgroups or highly uncertain subgroups. We are interested in such summarization on the entire dataset to see the whole uncertainty level, and on the subgroups to see how the descriptive information interplays with the uncertainty information. We use these measures to define a quality measure for the mSMoPE model class, that gauges how exceptional the uncertainty set is on a subgroup compared to the uncertainty set on the entire dataset. 

\subsubsection{Generating $r$ for Classification Models} 
Suppose that the model that we are analyzing is a classification model and the dataset is a classification dataset. Firstly, we take the softmax output of the model as the soft outputs. Then we can set the conformal score $s_i = 1 - \hat{\mu}(X_i)_{Y_i}$ to be one minus the softmax output of the true class. Because we do not know whether the probability of softmax output is any good, we can only treat it as the heuristic notion of the uncertainty. Secondly, we define $\hat{q}$ to be the $[(n+1)(1-\alpha)]/n$ empirical quantiles of $\{s_1, \ldots, s_n\}$ on $\Omega_{\text{calib}}$. Thirdly, for $\Omega_{\text{test}}$, we can create a prediction set $\mathcal{T}(X_{\text{test}}) = \{y : \hat{\mu}(X_{\text{test}}) \geq 1 - \hat{q}\}$ that includes all classes with a high enough softmax output. Finally, for each record $(X_i,Y_i)\in\Omega_{\text{test}}$, we set $r=|T(X_i)|$: the size of its prediction set. The larger $r$ is, the less certain the model is of its prediction.

\subsubsection{Generating $r$ for Regression Models}
Suppose that the model we are analyzing is a regression model and the dataset is a regression dataset. Firstly, we use the quantile regression model to get an initialized interval as $[\hat{t}_{\alpha/2}(x), \hat{t}_{1 - \alpha/2}(x)]$. With the quantile output, we can define the score function to be the projective distance from $y$ onto the interval as: $s(x, y) = \max \{\hat{t}_{\alpha/2}(x) - y, y - \hat{t}_{1 - \alpha/2}(x)\}$. Here the scores are computed using $\Omega_{\text{calib}}$. Secondly, we compute $\hat{q} = \text{Quantile}(s_1,\ldots,s_n; [(n+1)(1-\alpha)]/n)$ and formulate the valid prediction set by taking $\mathcal{T}(x) = [\hat{t}_{\alpha / 2}(x) - \hat{q}, \hat{t}_{1-\alpha/2}(x) + \hat{q}]$. Intuitively, the set $\mathcal{T}$ just grows or shrinks the distance between the quantiles by $\hat{q}$ to achieve coverage. Finally, for each record $(X_i,Y_i)\in\Omega_{\text{test}}$, we set $r$ equal to the length of the interval for $\mathcal{T}(X_i)$. The larger $r$ is, the more difficult the model finds the input data.

\subsubsection{Quality Measures over $r$}
For the entire dataset or a given subgroup, we have a distribution of target variable $r$, which represents the distribution of uncertainty on the population level. In practice, we extract the average value of the distribution as the representation of the target variable, denoted as Average Uncertainty Loss.

\begin{definition}[Average Uncertainty Loss]
Given a (sub-)population $S$ of records in a dataset, the average uncertainty loss (AUL) is given by:
\begin{equation*}
\text{AUL}(S) = \frac1{|S|}{\sum_{(X_i,Y_i)\in\ S} r^i}
\end{equation*}
\end{definition}
Here, $r^i$ represents the computed size of prediction set or the length of interval for record $i$. 

In Exceptional Model Mining, we strive to find subgroups for which the target interaction captured by the model class is exceptional. Exceptionality does not occur in a vacuum: the behavior on a subgroup can only be exceptional when contrasted with a reference behavior that represents normality. Often, target behavior across the full dataset is used for this reference behavior, and so we define the following quality measure for the mSMoPE model class.

\begin{definition}[Relative Average Uncertainty Loss]
Given a subgroup $S$ of $\Omega$, its Relative Average Uncertainty Loss, $\varphi_{\text{raul}}$, is given by:
\begin{equation*}
\varphi_{\text{raul}}(S) = \text{AUL}(\Omega) - \text{AUL}(S)
\end{equation*}
\end{definition}
To find subgroups for which the model is highly certain about its prediction, i.e., subgroups for which the soft model works very well, one should maximize $\varphi_{\text{raul}}$; positive values for $\varphi_{\text{raul}}$ indicate that the soft model performs better than usual on this subgroup. It is because the efficiency of conformal prediction is higher than usual. To find subgroups for which the soft model does not work, one should minimize $\varphi_{\text{raul}}$; negative values for $\varphi_{\text{raul}}$ indicate that the soft model performs worse than usual on this subgroup. It is because the efficiency of the conformal prediction is lower than usual.
Alternatively, one could find a list of subgroups for which the soft model performs exceptionally in general, by maximizing $|\varphi_{\text{raul}}|$. The resulting list of subgroups could be partitioned into poorly- and well-classified subgroups in a post-processing step. In this paper, however, we maintain the strict separation of bad and good subgroups by presenting results of $\varphi_{\text{raul}}$-maximizing and -minimizing runs separately.

The definitions of these quality measures themselves are trivial. The main contributions of Conformalized EMM and the mSMoPE model  class for EMM lie not so much in convoluted quality measure formulas, but instead in the path taken through Conformal Prediction methodologies in order to arrive at meaningful definitions of $r$ (to subsequently be incorporated in quality measures for EMM).

\section{Experimental Setup}

We evaluate the performance of Conformalized EMM in several experiments, performed on several real-world datasets, and demonstrating the exceptional subgroups Conformalized EMM can discover. By doing this, we aim to show the effective performance through the newly defined model class mSMoPE and the associated quality measure $\varphi_{\text{raul}}$. The contents are organized as follows. Firstly, we introduce the datasets used in this paper. We introduce the source of the data, the detailed composition of the data and the tasks in the data. Secondly, we introduce the base model $\mu$ that is required to conduct Conformalized EMM. Thirdly, we introduce the results, by listing the subgroups discovered in the datasets. Source code for reproducing the experiments are released \url{https://github.com/octeufer/ConformEMM}.
%
%
Through the experiments, we aim to answer the following questions:

\begin{enumerate}
  \item Is Conformalized Exceptional Model Mining sufficient to tell where your model performs (not) well?
  \item Can Conformalized Exceptional Model Mining efficiently discover meaningful subgroups from multiple datasets?
  \item How does the performance of Conformalized Exceptional Model Mining vary across models and datasets?
\end{enumerate}

\subsection{Datasets}
We use a diverse set of public datasets. These datasets are: \textbf{Wine quality}, for which the goal is to model wine quality based on physicochemical tests~\cite{asuncion2007uci}. \textbf{Online News Popularity}, which summarizes a heterogeneous set of features about the articles published by Mashable in a period of two years. The goal is to predict the level of shares in social networks~\cite{asuncion2007uci}. \textbf{Helena}, an anonymized dataset for the classification task, including 100 classes~\cite{guyon2019analysis}. \textbf{Covertype}, classification into forest cover types based on characteristics such as elevation, aspect, slope, hillshade, and soil type~\cite{blackard1999comparative}. \textbf{MimicIII}, a large database comprising deidentified health-related data associated with over forty thousand patients who stayed in critical care units of the Beth Israel Deaconess Medical Center~\cite{johnson2016mimic}. \textbf{California Housing}, real estate data~\cite{pace1997sparse}. \textbf{Year, the Million Song Dataset}, a collection of audio features and metadata for a million contemporary popular music tracks~\cite{Bertin-Mahieux2011}. Table~\ref{tb:datasets} lists metadata of all these datasets.

\begin{table}[t]
\centering
\caption{Dataset metadata, with the average uncertainty loss for the MLP model.}
\label{tb:datasets}
\begin{tabular}{llrrlr}\toprule
 $i$   & $\Omega_i$  & $N$ & $k$ & Task & AUL($\Omega_i$) \\ 
\midrule
1 & Wine Quality & 4\thinspace898 & 11 & classification & $2.471$ \\
2 & News Popularity & 39\thinspace644 & 59 & classification & $8.377$  \\
3 & Helena & 65\thinspace196 & 27 & classification & $33.277$ \\
4 & Cover Type & 581\thinspace012 & 54 & classification & $1.085$  \\
5 & MimicIII & 7\thinspace414 & 19 &  regression & $319.305$  \\
6 & CA Housing & 20\thinspace640 & 8 & regression & $279\thinspace400.923$   \\
7 & Year & 515\thinspace345 & 90 & regression & $25.161$  \\
\bottomrule
\end{tabular}
\end{table}

\subsection{Setting Up the Predictor}
Conformalized EMM requires a predictor function $\mu$, whose conformal predictions lead to the generation of the EMM target space $r$, which ought to represent the uncertainty regarding the performance of the model (for classification or regression tasks). We employ a Multi-Layer Perceptron (MLP) as this base model. The MLP that we implement follows common design principles. We compose the model from multiple MLP blocks, each constructed by a linear fully-connected layer with batch normalization~\cite{ioffe2015batch}, ReLU~\cite{arora2018understanding}, and dropout~\cite{srivastava2014dropout}. We select this NN architecture because it can handle all types of available attributes (binary, nominal, numeric). Moreover, the architecture is flexible in the design so that it can handle softmax outputs and quantile regression outputs. 

\subsection{Parameterization}

As outlined in Section~\ref{sec:msmope}, we implement the CP algorithms so that the size of the prediction set and the length of the prediction interval can be computed based on the MLP model. The mSMoPE model class and the $\varphi_{\text{raul}}$ quality measure are implemented following~\cite{du2020fairness,du2021beyond}, where we restrict the search to a refinement depth of 2, i.e., we allow the resulting subgroups to be defined on two condition of the descriptors. This setting explores the expressive power of the resulting subgroups and enhances their potential for the interpretation by domain experts. 
Nothing restricts the use of the mSMoPE model class to subgroups defined by only two attributes. We opt for this limitation to maintain a balance between expressiveness and interpretability: exploring subgroups with more attributes would be computationally feasible at the cost of less-interpretable results.
The search space is defined based on the types of attributes. For a binary attribute $a_i$, we consider the subgroups $a_i=0$ and $a_i=1$. If $a_i$ is a nominal attribute with \emph{m} distinct values $v_1, \ldots, v_m$, we examine \emph{m} subgroups of the form $a_i = v_j$. For a real-valued attribute, we explore multiple intervals using the dataset's observed values as interval endpoints. Among these, only two subgroups are reported: the highest-scoring subgroup of the form $a_j \leq v_j$ and the highest-scoring subgroup of the form $a_i > v_j$.
Finally, we use a parameter $\lambda$ to bound the minimum subgroup size, to combat overfitting: we only report subgroups that contain at least $\lambda\%$ of the records in the dataset.

We run our algorithms twice for each dataset: once maximizing $\varphi_{\text{raul}}$ in order to find subgroups on which the model performs well, with high certainty about its outputs, and once minimizing $\varphi_{\text{raul}}$ in order to find subgroups on which the model performs poorly, with high uncertainty. In each run, we only report subgroups whose AUL outperforms the baseline set by the AUL of the entire dataset: the maximizing run reports only subgroups with $\varphi_{\text{raul}} \geq 0$, and the minimizing run reports only subgroups with $\varphi_{\text{raul}} \leq 0$.

\section{Experimental Results}

\begin{table}[t]
\centering
\caption{Top-3 subgroups per dataset maximizing $\varphi_{\text{raul}}$.}
\label{tb:spdm01}
\begin{tabular}{llr}
\toprule
 $\mathbf{\Omega_i}$   & \textbf{Most certain subgroups} $\mathbf{S}$ & $\mathbf{\varphi_{\text{raul}}(S)}$ \\ 
\midrule
\multirow{3}{*}{$\Omega_1$~~~~} & free sulfur dioxide $> 130.444$ and alcohol $> 10.756$ & $1.471$ \\
&free sulfur dioxide $> 130.444$ and sulphates $\leq 0.507$ & $1.471$ \\
&free sulfur dioxide $> 130.444$ and sulphates $\leq 0.411$ & $1.471$ \\
\midrule
\multirow{6}{*}{$\Omega_2$} & self reference max shares $> 562\thinspace200.000$ & \multirow{2}{*}{$7.082$} \\
&~~~~~~~~and self reference avg shares $\leq 76\thinspace711.111$\\
& self reference max shares $> 749\thinspace600.000$  & \multirow{2}{*}{$6.532$} \\
&~~~~~~~~and min negative polarity $\leq -0.667$\\
&self reference max shares $> 562\thinspace200.000$ & \multirow{2}{*}{$6.282$} \\
&~~~~~~~~and num self hrefs $> 25.778$\\
\midrule
\multirow{3}{*}{$\Omega_3$} & V12 $> 226.693$ and V10 $\leq 170.078$ & $30.025$ \\
&V21 $\leq -14.767$ and V12 $> 198.388$ & $29.809$ \\
&V21 $\leq -14.767$ and V19 $> 77.927$ & $29.612$ \\
\midrule
\multirow{6}{*}{$\Omega_4$} & Hillshade Noon $\leq 56.444$ & \multirow{2}{*}{$0.085$} \\
&~~~~~~~~and Horizontal Distance To Fire Points $\leq 7\thinspace173.000$~~~~\\
&Hillshade Noon $\leq 56.444$  & \multirow{2}{*}{$0.085$} \\
&~~~~~~~~and Horizontal Distance To Fire Points $\leq 6\thinspace376.000$\\
&Hillshade Noon $\leq 56.444$  & \multirow{2}{*}{$0.085$} \\
&~~~~~~~~and Horizontal Distance To Fire Points $\leq 5\thinspace579.000$\\
\midrule
\multirow{3}{*}{$\Omega_5$} & GLUCOSE $> 538.333$ and SODIUM $\leq 135.667$ & $136.275$ \\
&GLUCOSE $> 538.333$ and ALBUMIN $> 3.133$ & $133.106$ \\
&GLUCOSE $> 538.333$ and CHLORIDE $\leq 106.000$ & $132.774$ \\
\midrule
\multirow{3}{*}{$\Omega_6$} & population $> 3\thinspace967.333$ and households $\leq 1\thinspace352.333$ & $51\thinspace232.926$ \\
&population $> 3\thinspace967.333$ and total rooms $\leq 8\thinspace432.000$ & $51\thinspace137.378$ \\
&longitude $> -115.426$ and median income $\leq 2.111$ & $49\thinspace965.760$ \\
\midrule
\multirow{3}{*}{$\Omega_7$} & V35 $\leq -342.419$ and V6 $\leq -1.723$ & $10.106$ \\
&V35 $\leq -342.419$ and V1 $> 41.529$ & $9.671$ \\
&V3 $\leq -94.950$ and V35 $\leq -201.227$ & $9.655$ \\
\bottomrule
\end{tabular}
\end{table}

\begin{table}[t]
\centering
\caption{Top-3 subgroups per dataset minimizing $\varphi_{\text{raul}}$.}
\label{tb:spdm02}
\begin{tabular}{llr}
\toprule
 $\mathbf{\Omega_i}$   & \textbf{Most uncertain subgroups} $\mathbf{S}$ & $\mathbf{\varphi_{\text{raul}}(S)}$ \\ 
\midrule
\multirow{3}{*}{$\Omega_1$} & sulphates $> 0.984$ and total sulfur dioxide $> 152.667$ & $-2.529$ \\
&sulphates $> 0.984$ and free sulfur dioxide $\leq 33.889$ & $-1.529$ \\
&chlorides $> 0.271$ and sulphates $\leq 0.507$ & $-1.529$ \\
\midrule
\multirow{3}{*}{$\Omega_2$} & LDA 04 $> 0.927$ and abs title sentiment polarity $\leq 1.000$ & $-1.622$ \\
&LDA 04 $> 0.927$ and abs title sentiment polarity $> 0.333$ & $-1.622$ \\
&LDA 04 $> 0.927$ and abs title sentiment polarity $> 0.222$ & $-1.622$ \\
\midrule
\multirow{3}{*}{$\Omega_3$} & V11 $\leq 28.576$ and V2 $\leq 0.342$ & $-17.820$ \\
&V11 $\leq 28.576$ and V6 $\leq 0.371$ & $-17.276$ \\
&V2 $\leq 0.122$ and V10 $\leq 56.849$ & $-17.271$ \\
\midrule
\multirow{6}{*}{$\Omega_4$} & Horizontal Distance To Fire Points $> 6\thinspace376.000$ & \multirow{2}{*}{$-0.790$}\\
&~~~~~~~~and Hillshade 3pm $> 225.778$\\
&Horizontal Distance To Fire Points $> 6\thinspace376.000$  & \multirow{2}{*}{$-0.790$} \\
&~~~~~~~~and Hillshade 9am $\leq 112.889$\\
&Elevation $\leq 2\thinspace525.333$  & \multirow{2}{*}{$-0.665$} \\
&~~~~~~~~and Horizontal Distance To Fire Points $> 5\thinspace579.000$\\
\midrule
\multirow{3}{*}{$\Omega_5$} & BUN $> 87.667$ and ALBUMIN $\leq 2.600$ & $-114.185$ \\
&ALBUMIN $\leq 2.067$ and HEMOGLOBIN $> 10.867$ & $-108.213$ \\
&ALBUMIN $\leq 2.067$ and HEMATOCRIT $> 32.100$ & $-98.418$ \\
\midrule
\multirow{3}{*}{$\Omega_6$~~~~} & median income $> 6.944$ and housing median age $> 46.333$~~~~ & $-117\thinspace202.286$ \\
&median income $> 6.944$ and housing median age $> 40.667$ & $-115\thinspace196.680$ \\
&median income $> 8.556$ and housing median age $> 46.333$ & $-114\thinspace451.983$ \\
\midrule
\multirow{3}{*}{$\Omega_7$} & V6 $> 8.905$ and V2 $\leq -164.854$ & $-18.338$ \\
&V3 $> 214.741$ and V35 $\leq 118.308$ & $-17.146$ \\
&V3 $> 214.741$ and V24 $\leq 1054.198$ & $-17.146$ \\
\bottomrule
\end{tabular}
\end{table}

For each dataset, the top-3 most certain subgroups found while maximizing $\varphi_{\text{raul}}$ are reported in Table~\ref{tb:spdm01}, along with their qualities. When comparing the final columns of Tables~\ref{tb:datasets} and~\ref{tb:spdm01}, the values for the datasets Wine Quality ($\Omega_1$) and Cover Type ($\Omega_4$) stand out. For the subgroups found on these datasets, we see that $\varphi_{\text{raul}}$ is smaller than AUL($\Omega_i$) by precisely 1, which implies that the average uncertainty loss of the top-ranked subgroups is just less than the average uncertainty loss on the whole dataset. I.e.: Conformalized EMM discovered subgroups for which the classifiers are highly confident about their prediction set so that the label is just the only one in the set. Even for datasets News Popularity and Helena, the size of prediction set in the subgroups that we found is limited to 2 and 3 in average uncertainty loss. We can see that the average uncertainty loss on the whole datasets for these datasets are $8.377$ and $33.277$. It shows that Conformalized EMM discovers subgroups that are extremely certain about their predictions comparing with the average uncertainty loss on the entire datasets. For the regression tasks, we see that our algorithms found subgroups that decrease the uncertainty prediction interval from $319.305$ by $136.275$, from $279\thinspace400.923$ by $51\thinspace232.926$, and from $25.161$ by $10.106$. This is a substantial drop. More results and visualizations of our findings can be found in the supplementary material~\cite{Du2025}.

On the other hand, when comparing the final columns of Tables~\ref{tb:datasets} and~\ref{tb:spdm02}, the values for the datasets Wine Quality ($\Omega_1$) and News Popularity ($\Omega_2$) stand out. For the subgroups found on these datasets, we see that $\varphi_{\text{raul}}$ is smaller than AUL($\Omega_i$), which implies that the average uncertainty loss of the top-ranked subgroups is just larger than average uncertainty loss on the entire dataset. I.e.: Conformalized EMM discovered subgroups for which the classifiers are highly unconfident about their prediction set. We notice that for News Popularity dataset, the discovered subgroups demonstrate an uncertainty set of size 10, which is the largest set we encountered within our experiments: Conformalized EMM identified the prediction that contains all the possible labels in the prediction set, which shows the most uncertain performance.




\section{Conclusion}
We introduce Conformalized Exceptional Model Mining (Conformalized EMM): a method to discover subgroups in a dataset where a multi-class classifier or regressor is exceptionally certain or uncertain about its own prediction. We express this (un)certainty in terms of the size of the Adaptive Prediction Set for multi-class classification problems, and the interval length in Conformalized Quantile Regression for regression problems. The (un)certainty expression is subsequently fed to the mSMoPE (multiplex Soft Model Performance Evaluation) model class for EMM, discovering exceptional subgroups evaluated by the Average Uncertainty Loss, a quantity expressing how well the soft model outputs can represent the confidence of the model predictions. The quality measure $\varphi_{\text{raul}}$ is designed to find coherent subspaces of the dataset where the soft model performs highly certain (when maximizing $\varphi_{\text{raul}}$), highly uncertain (when minimizing $\varphi_{\text{raul}}$) or exceptional (when maximizing $|\varphi_{\text{raul}}|$). Since EMM results in easily interpretable subgroups, our focus is not on letting the machine improve the machine: the primary goal in the mSMoPE model class for EMM is to provide a better understanding to the domain expert. 

We illustrate the findings one could expect from the mSMoPE model class by experiments on seven datasets. Some discovered subgroups are troublesome for our model, and some subgroups are found where our model has barely any problems. The mSMoPE model class highlights as a particularly troublesome area a subgroup in News Popularity dataset whose characterizing feature is associated with all the classes in the dataset, and as a particularly benign area a subgroup that is associated with one particular class. Overall, when maximizing $\varphi_{\text{raul}}$, one easily finds small subgroups on which the soft model performs highly certain; the subgroups on which the soft model performs highly uncertain are typically less trivial, hence they demand further attention. The mSMoPE model class for EMM helps to understand multi-class classification and regression, which leads to ideas on how to improve the overall classifier or regressor performance. 

\begin{credits}
\subsubsection{\ackname} This work was supported by National Natural Science Foundation of China (NSFC) under Grant No.62476047. This work was partly supported by H2020 SmartChange project, funded within EU’s Horizon Europe research program (GA No. 101080965) and NWO EDIC project.

\subsubsection{\discintname}
The authors have no competing interests to declare that are relevant to the content of this article.
\end{credits}




\end{document}